\documentclass{article}

\usepackage{arxiv}

\usepackage[utf8]{inputenc} % allow utf-8 input
\usepackage[T1]{fontenc}    % use 8-bit T1 fonts
\usepackage{hyperref}       % hyperlinks
\usepackage{url}            % simple URL typesetting
\usepackage{booktabs}       % professional-quality tables
\usepackage{amsfonts}       % blackboard math symbols
\usepackage{nicefrac}       % compact symbols for 1/2, etc.
\usepackage{microtype}      % microtypography
\usepackage{lipsum}\usepackage{amsfonts}   % blackboard math symbols
\usepackage{nicefrac}       % compact symbols for 1/2, etc.
\usepackage{microtype}      % microtypography
\usepackage{lipsum}
\usepackage{graphicx}
\usepackage{amsmath}
\usepackage{pbox}
\usepackage[usestackEOL]{stackengine}
\usepackage{longtable}
\usepackage{caption}
\usepackage[round]{natbib}
\usepackage{hyperref}
\usepackage{caption}
\usepackage{subcaption}
\usepackage{amsmath}
\usepackage{MnSymbol}
\usepackage{wasysym}
\usepackage{arabtex}
\usepackage{utf8}
\usepackage{cleveref}

\title{Evaluating Various Tokenizers for Arabic Text Classification}

\author{
  Zaid Alyafeai\\
  Department of Computer Science\\
  King Fahd University of Petroleum and Minerals\\
  Dhahran, Saudi Arabia 31261 \\
  \texttt{g201080740@kfupm.edu.sa} \\
  \And
  Maged S. Al-shaibani \\
  Department of Computer Science\\
  King Fahd University of Petroleum and Minerals\\
  Dhahran, Saudi Arabia 31261 \\
  \texttt{g201381710@kfupm.edu.sa} \\
  \And
  Mustafa Ghaleb \\
  Department of Computer Science\\
  King Fahd University of Petroleum and Minerals\\
  Dhahran, Saudi Arabia 31261 \\
  \texttt{g200905270@kfupm.edu.sa} \\
  \And
  Irfan Ahmad \\
  Department of Computer Science\\
  King Fahd University of Petroleum and Minerals\\
  Dhahran, Saudi Arabia 31261 \\
  \texttt{irfan.ahmad@kfupm.edu.sa} 
   
}

\begin{document}
\setcode{utf8}
\maketitle

% \pagebreak

\begin{abstract}
The first step in any NLP pipeline is to split the text into individual tokens. The most obvious and straightforward approach is to use words as tokens. However, given a large text corpus, representing all the words is not efficient in terms of vocabulary size. In the literature, many tokenization algorithms have emerged to tackle this problem by creating subwords which in turn limits the vocabulary size in a given text corpus. Most tokenization techniques are language-agnostic i.e they don't incorporate the linguistic features of a given language. Not to mention the difficulty of evaluating such techniques in practice. In this paper, we introduce three new tokenization algorithms for Arabic and compare them to three other baselines using unsupervised evaluations. In addition to that, we compare all the six algorithms by evaluating them on three supervised classification tasks which are sentiment analysis, news classification and poetry classification using six publicly available datasets. Our experiments show that none of the tokenization technique is the best choice overall and that the performance of a given tokenization algorithm depends on the size of the dataset, type of the task, and the amount of morphology that exists in the dataset. However, some tokenization techniques are  better overall as compared to others on various text classification tasks.

\end{abstract}

\section{Introduction}

Tokenization is the process of breaking text into a list of tokens. These tokens are encoded using integers then fed into machine learning models. One possible way is to split text into words which have intrinsic meaning and white-spaces can easily be utilized for tokenization. However, using words as tokens comes with its own problems like the vocabulary size that can grow with the size of the training dataset. A typical solution to this problem is to reduce the vocabulary size but this will result in a lot of unknowns especially for languages that have rich morphology. On the other hand, character tokenization can result in a smaller vocabulary size and reduced unknowns but they usually require a huge number of parameters to model \cite{al2019character}. Fine-grained tokenization like character tokenization, generally, requires deeper and wider models while coarse-grained tokenization like word tokenization usually requires larger embeddings to learn reasonable representations.  A good tokenization approach achieves a balanced granularity between characters and white space segregated tokens. Subword tokenizations allows splitting words into units called subwords. The ability of creating subwords allows designing universal tokenization across different languages in addition to limiting the vocabulary size. Theoretically, we can think of subword tokenization as a mid-level granularity between character and word tokenization. Still, subword tokenization comes with its own problems like creating a lot non-meaningful subwords. 

In the literature, tokenization has gained a lot of attention especially with the rise of the recent transformer-based architectures to limit the vocabulary size especially for large datasets. In addition to that, resulting subword units can be combined to construct new vocabulary that does not appear in the training corpus which allows generalizing to new domains or datasets via fine-tuning. Such tokenizations are widely adopted in many recently proposed transformer-based architectures such as BERT \citep{devlin2018bert}, T5 \citep{raffel2019exploring} and GPT-2 \citep{radford2019language}. The most prominent tokenizers in the literature are byte pair encoding BPE \citep{sennrich2015neural}, WordPiece \citep{schuster2012japanese}, SentecePiece \cite{kudo2018sentencepiece}, Huffman encoding \citep{chitnis2015variable}, and orthographic syllable \citep{kunchukuttan2016orthographic}. 

Data-driven tokenizations like Byte Pair Encoding (BPE) generally provide multiple levels of granularity. This granularity, in the case of frequency-based approaches, is controlled by tokens' frequency in the training corpus. Most frequent tokens are kept intact while the least frequent ones are broken into their frequent subword pieces. In extreme cases, it will end up broken into its constituent characters. Other approaches, like Morfessor \citep{SmitVirpiojaGronroosEtAl2014}, devised an unsupervised language-agnostic tokenization. The tokenization model in their approach learns to distinguish morphemes from their affixes by analyzing the corpus for language patterns. The level of granularity in this approach is influenced by language morphology. 

Many tokenization approaches can be devised for a given language. Although plain frequency-based tokenizations are dominating, they are morphologically blind. For example, some researchers report limitations faced of subword tokenization like BPE as it results in  sub-optimal segmentation in cross-lingual settings \citep{wang2021multi}. Other directions can be explored to glue this sub-optimality gap like introducing a morphology aware approach. Morphological tokenization can take advantage of the language knowledge it is shipped with. Hence, there is a good chance they may compete with the BPE tokenizations especially for a morphologically rich languages like Arabic. 

Arabic is a highly inflected language with rich morphology. There are two types of morphology which are derivational and inflectional. Inflectional morphology is used to inflect gender, tense, etc. from a given stem. For example the stem \< 
ذهب
> (go) could have the present tense inflection \< يذهب > (to go) the plural inflection \<يذهبون> (they go) and the feminine inflection \< ذهبت > (she went) . Derivational morphology on the other hand causes changes to the meaning of a word or part of speech. For instance, given the stem \<
كتب
>
(write) could have the derivations \< كاتب > (writer), \< كتابة > (writing) and \<مكتبة > (library). As we saw from the inflections the added morphemes could be at the beginning (affixes), at the middle (infixes) or at the end (suffixes). The cursive nature of Arabic makes it difficult to recognize the stem from the added morphemes. Such features challenge tokenization in many tasks.  Also, due to its morphological density, it is important to evaluate the performance of different types of tokenization scheme using a variety of tasks.  

With the variety of tokenization approaches in hand, given that the tokenization is task sensitive, it becomes difficult to choose the best approach for a given task. Additionally, tokenization evaluation is not a straightforward process. A reasonable way to evaluate the tokenization performance on a given task is to evaluate the output produced by the model with the specific tokenization on that task. We call such evaluation, in this study, as supervised evaluation. Moreover, other attributes can be considered. For example, the tokens text coverage resulting from the tokenization step and the tokenization speed. We call such type of evaluation as unsupervised evaluation.

In this research, we aim at evaluating the performance of six different tokenization schemes on three different tasks, sentiment analysis, news classification, and poem-meter classification. We summarize our contributions as follows: 

\begin{enumerate}
    \item We introduce three new tokenization techniques for Arabic which are: Stochastic tokenizer, Disjoint-letter tokenizer and Morphological tokenizer. 
    \item We design two types of evaluations based on supervised and unsupervised approaches.  
    \item We evaluate the six different tokenizers on three different tasks with different vocabulary sizes. 
\end{enumerate}

The rest of this paper is structured as follows. In \textbf{Section 2} we discuss the existing literature on this topic. In \textbf{Section 3} we present the six tokenizers that are evaluated in this work. In \textbf{Section 4}, we discuss the tasks and the datasets that are used for evaluating the tokenizers. The experiment setup and discussion of the model architectures are outlined in \textbf{Section 5}. For completeness purposes, in section \textbf{Section 6} we introduce two unsupervised evaluation measures for tokenizers: speed and compression factor. In \textbf{Section 7}, we present the results of the used datasets in terms of accuracy in 6 different vocabulary sizes. 
Finally, \textbf{Section 8} concludes the findings and envisions future directions.

\section{Related Work}

In the past few years, tokenizations went through many stages. The earlier stages used word level tokenization with pretrained embeddings using Word2Vec \cite{mikolov2013efficient} and GloVe. \cite{pennington2014glove}. Later, n-gram embeddings have gained some attention using fasttext \cite{bojanowski2017enriching}. Eventually, subword tokenization algorithms have gained a lot of interest, mainly using BPE algorithm \cite{sennrich2015neural}, WordPiece \cite{wu2016google} and unigram language modelling  \cite{kudo2018subword}. Subword tokenization have been used in many popular models like BERT \cite{devlin2018bert} , RoBERTA \cite{liu2019roberta} and GPT-2 \cite{radford2019language}. SentencePiece \cite{kudo2018sentencepiece} have also been suggested as a modification of these algorithms to process raw data instead of tokenized words. This is important especially in languages that don't have word boundaries like Chinese. Recently free-tokenization approaches have picked up a lot of interest. Mainly, character-based models like \cite{al2019character} and Byte-based models \cite{xue2021byt5} are interesting but they usually require very deep architectures. Recent advances suggest creating end-to-end systems to tokenize on the Byte level and encoding some word boundaries \cite{tay2021charformer}. 

In Arabic NLP, subword tokenization is widely used in many tasks in the recent literature. HULMonA \citep{eljundi2019hulmona} used the WordPiece tokenization approach for the task of language modeling for Arabic based on UMLFiT architecture \cite{howard2018universal}. AraBERT  \citep{antoun2020arabert} used SentencePiece to train BERT model from scratch on Arabic corpus segmented using FARASA \cite{abdelali2016farasa}. They showed that AraBERT achieved advanced results on various tasks. Moreover, they show slight improvement over training without segmentation on various text classification tasks. Abdul-Mageed et al. \cite{abdul2020arbert} used WordPiece algorithm for training BERT-like model and evaluated it in many tasks. They show improvement over AraBERT in many tasks. Abdelali et al. \cite{abdelali2021pre} introduced QARiB a BERT-model trained on Arabic dialects. They show improvement in many datasets especially those that are collected from Twitter. They mainly employed BPE algorithm after a preprocessing step where they applied FARASA segmentation. 

In the literature, there are many studies that experiment with different tokenization schemes in different languages.  \citep{shapiro2018bpe} experimented and discussed the performance of charCNN and PBE tokenizations for translating from 8 different languages to English. \citep{sajjad2017challenging} also analyzed the performance of these two approaches in the context of Part of Speech tagging and Neural Machine Translation (NMT). Bostrom and Durret \cite{bostrom2020byte} showed that unigram language modelling is superior to BPE in retrieving more meaningful tokens. Moreover, when evaluated on Arabic and English they show that unigram can outperform BPE when fine-tuning on multiple tasks. In Arabic, \citep{shapiro2018morphological} studied the performance of subwords and words tokenization in the context of NMT from Arabic to English. \citep{alkaoud2020importance} studied the sensitivity of the tokenization approach by training SentencePiece on subwords and compare it to a variant of AraBERT tokenization where they gain slight improvements. Oudah et al. \cite{oudah2019impact} studied the effect of preprocessing Arabic text using segmentation on Arabic-English machine translation. They argue that the tokenization scheme depends on the data and the model used for training. Alkaoud and Syed \cite{alkaoud2020importance} showed the affect of using different tokenizations for fine-tuning BERT models. They show that character tokenization can outperform subword in poetry classification. 

In the context of the studied tasks in this research, the widely applied approaches for non-deep learning techniques in sentiment analysis and text classification are word-based tokenizations. These approaches split the text into tokens based on white spaces and non-alphabetical characters. Deep learning techniques, on the other hand, were not used due to the lack of large labeled datasets \citep{ABUFARHA2021102438}. The prime focus in these approaches is preprocessing steps and feature engineering techniques in order to improve the result. A set of recent surveys dedicated to this topic includes \citep{al2019comprehensive}, \citep{guellil2019arabic} and \citep{badaro2019survey}. In recent years, many studies have been applying different techniques for poetry meter classification. For example, \citep{al2020meter}, \citep{abandah2020classifying} and \citep{yousef2019learning} employed a character-based tokenization to split the poem verses into characters. It is interesting to analyse the behaviour of other subwords tokenizers on this task as the character tokenizations appear to perform the best.

As we see from the collected literature, there is no work that studies the affect of different tokenizers in Arabic. Moreover no language-specific tokenizers have been evaluated for Arabic. We attempt to fill this gap by introducing new language-specific tokenizers in addition to comparing them with other tokenizers using unsupervised and supervised evaluation techniques. 

\begin{figure}
    \centering
    \includegraphics[scale=0.3]{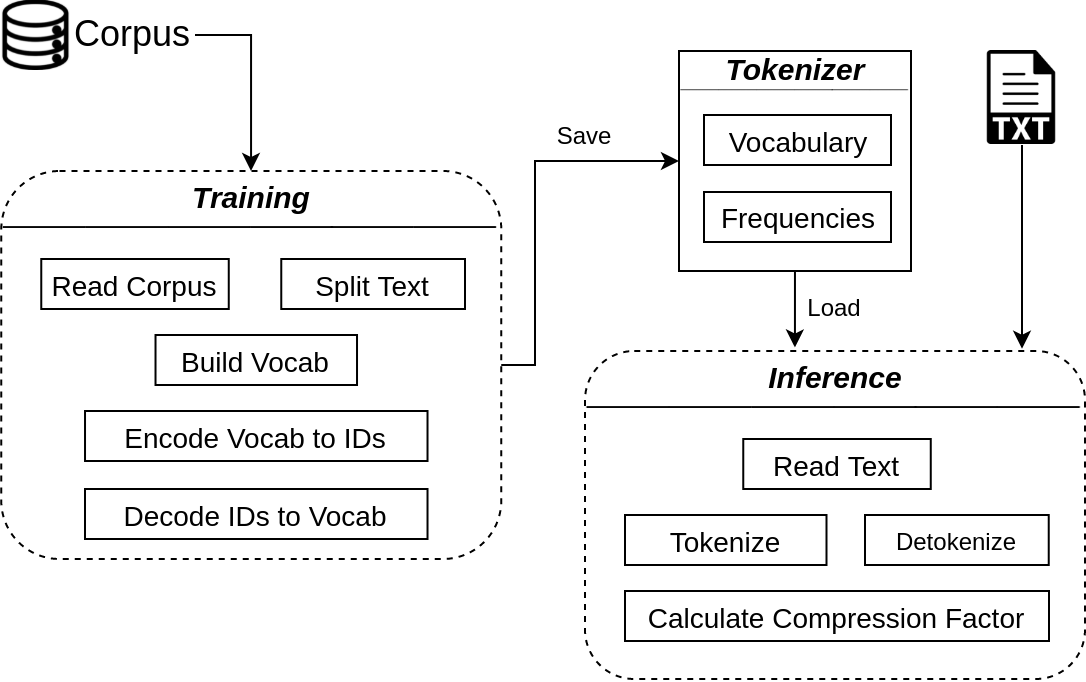}
    \caption{The Tokenization Framework.}
    \label{fig:tkseem}
\end{figure}

\section{Tokenization Schemes}
\label{section:tokenizers}
In this section, we illustrate how we built our tokenizer evaluation framework. The framework features the 6 tokenizers presented in this paper. It also contains many other auxiliary functions that facilitate the training and the inference processes as shown in Figure \ref{fig:tkseem}. Five of these tokenizers are built from scratch and the sixth one, SentencePiece, is integrated from the available library \footnote{\url{https://github.com/google/sentencepiece}}. Our framework is published as an open source python library on GitHub and PyPI. The source code is accessible through GitHub \footnote{\url{https://github.com/ARBML/tkseem}} along with the instructions to install and fine-tune on different tasks. 

\subsection{Tokenizers Description}

The list below introduces each tokenizer we used in this research. The new tokenizers that we introduce are Stochastic tokenizer, Morphological tokenizer and Disjoint letter tokenizer. 

\begin{enumerate}

    \item \textbf{Character tokenizer}
    
    As the name implies, words are tokenized into a sequence of characters. This is the most fine-grained tokenizer. For a large vocabulary size, this tokenizer will most likely cover all the characters in a given dataset. The implementation of the inference method comprises regular expressions to split words into characters. 
    
    \item \textbf{Word tokenizer}
    
    Just like the character tokenizer, this tokenizer splits text into its word tokens. The determination of tokens is controlled only by the white-space character. As in the case of character tokenizer, word tokenizer does not require any training except for calculating the most frequent words in the corpus. The use of a regular expression in this tokenizer boosts the inference process making it the fastest tokenizer as we will see in Section \cref{sec:unsupervised}. 
    
    \item \textbf{Morphological tokenizer}
    
    Arabic words comprise prefixes and suffixes coupled with the morpheme. The tokenizers in the literature doesn't use the linguistics features in the Arabic language. The most used approach in the literature is segmenting first then using other tokenizers like BPE \cite{antoun2020arabert}. Usually the segmentation process is slow especially if the corpus is very large. We design a fast tokenizer that can segregate affixes from their morphemes. In the training process, we first used MADAMIRA \citep{pasha2014madamira} morphological analyzer to segment words. The corpus used for training is AbulKhair corpus containing a large body of Arabic raw text of news \citep{el20161}. Then, based on the frequency, we saved these affixes along with the resulting base morphemes in the token-frequency pair dictionary as illustrated in the training section. 

    \item \textbf{Stochastic tokenizer}
    
      As we discussed in the Morphological tokenizer it tries to include some linguistic features in the tokenization process. On the other hand, we might argue that linguistic features are not important and any stochastic split to a given word will not affect the evaluation process.  
      In other words, we want to test if the meaningful split, like in the morphology tokenizer, has a noticeable effect on the applied NLP task.  Given a word that belongs to a training corpus, we choose a random number that represents the number of splits for that given word. For instance, when processing the word  \<
    بحر
    >
    we might get the random number 2 which says that all bigrams will be added to the training dictionary. 
    
    \item \textbf{Disjoint letter tokenizer}
    
    This tokenizer is inspired by the cursive nature of the Arabic script. In the Arabic writing system, a letter in a word may or may not join its successor letter. To the best of our knowledge, this tokenizer was not evaluated on the three NLP tasks experimented in this paper, poetry meter classification, sentiment analysis, and news classification. However, it is a well-known segmentation technique applied on images and is mostly used in Arabic Optical Characters Recognition. The image is scanned to identify the text, the text is then segmented into words, the words are, then, split by this tokenizer and the resulting subwords are further split into individual letters. These researches \citep{jasimarabic},\citep{atallah2009comparative}
     and \citep{al2017arabic} provide detailed discussions on this subject. Regarding the training process, we split words into subwords based on their constituting letter types, either joint or disjoint. Then we save the resulting subwords along with their frequencies in a dictionary.

    \item \textbf{SentencePiece tokenizer}
    
    SentencePiece is an unsupervised data-driven approach for subword tokenization. It treats the training corpus as a sequence of raw Unicode characters i.e there is no pre-tokenization step where words are extracted based on the white-space character.  This makes the algorithm language-agnostic because it can be applied to any sequence of characters. The default algorithm used in the tokenization process is BPE.  In our framework, we provide an API wrapping the well-known language-agnostic frequency-based SentencePiece tokenizer \citep{kudo2018sentencepiece}. The motivation behind building this wrapper is to facilitate any future research where multiple tokenizers including SentencePiece are utilized. SentencePiece attracts a lot of attention in the recent research and has been used in many deep learning architectures. It is considered the standard de facto tokenizer.
    
\end{enumerate}

In Table \ref{tab:compar} we compare all the tokenizers in terms of different metrics. Coverage: indicates if the tokenizers covers all the text and does not lead to OOVs (out of vocabulary words). Morphology: language properties are required. Randomness: stochastic procedure introduced. Language-agnostic: doesn't depend on the language. Granularity: describes if the tokenizers causes shorter or longer tokens.

\begin{table}[!htp]
\caption{Comparing the features of all tokenizers. }
\label{tab:compar}
\begin{tabular}{p{3.0cm}|c|c|c|c|c|c}
\hline
\textbf{Metric} & \textbf{Character} & \textbf{Word} & \textbf{Morphological} & \textbf{Stochastic} & \textbf{Disjoint letter} & \textbf{SentencePiece}  \\ \hline
Coverage & $\checkmark$ &   &   &      &     & $\checkmark$       \\ \hline
Morphology   &     &   &   $\checkmark$  &    & $\checkmark$    &     \\ \hline 
Randomness   &     &   &     &  $\checkmark$   &     &     \\ \hline 
language-agnostic & $\checkmark$      & $\checkmark$   &     &  $\checkmark$   &     & $\checkmark$    \\ \hline 
Granularity level &  High    & Low   & Low   & Medium  & Medium    & Medium    \\ \hline 
\end{tabular}
\end{table}

\subsection{Training}

The input to this module is a text corpus and the output is a dictionary of key-value pairs holding the tokens along with their frequencies referred to as tokenizer's vocabulary.  The values in the dictionary are the tokens frequency scores learned from the input corpus. The size of the vocabulary is a hyperparameter that is tuned according to the task and the model architecture. In our framework, we permit users to set this hyperparameter. Moreover, we allow  adding special tokens to this dictionary such as padding and \textit{unknown} tokens. For example, we use a special token like <unk> to represent unknown words. To avoid any confusion later in the inference, it is recommended to encode these special tokens with Unicode characters or special strings that do not usually appear in normal text. Furthermore, to enhance the process of building this dictionary, we implemented a custom memory mapping function that reads the text stream into memory in parallel. We utilized such method because it is expected that the corpus files can be very large. Hence, the use of the built-in Python functions for reading files will make the process extremely slow. A further optimization that is worth mentioning is the ability to save and load the trained dictionary. This feature enhances the tokenization portability so that the user trains a dictionary only once, then loads it on demand.

\subsection{Inference}

The inference task is responsible for breaking a given text into tokens based on a given tokenizer. Usually, these tokens are encoded into integer ids. The encoding and the decoding methods have been implemented in the framework for that purpose. If a word is broken into multiple subwords, we identify these subwords by a special symbol, '\#\#' at the beginning of the subword. This is the same symbol used by BERT WordPiece algorithm \cite{devlin2018bert}. However, for presentation purposes in this paper, we replaced these '\#\#' by '+' in our examples, \ref{tab:tokenization_examples}. The framework also implements a detokenization method for each tokenizer. This method takes a tokenized string and returns the detokenized format of that string. To give an example of the tokenization process, consider the Arabic word 
\<
الجمال
>. Following a frequency-based tokenization for this word, we first break it into all possible unique splits all the way to its constituent characters. Accordingly, the number of possibilities grows exponentially with the word length. As this process can take a long time for longer words, a maximum size flag is required. We set this flag to 20 characters. However, a user can change the value of this flag depending on the need. We speed up the split process following a dynamic programming approach. We cache the previous word splits in memory and use them on-demand instead of making redundant splits.  Once all possible splits are returned, we select the valid subwords splits by consulting the tokenization dictionary learned during the training task. If multiple splits are available, we favor the one with the highest frequency score, i.e., the most frequent subwords splits are returned as an output. A possible output for that word using the morphological tokenization is \<ال> and \<+جمال>. In table \ref{tab:tokenization_examples} we present the output of each tokenization when given the following full Arabic sentence: \<
 سابح في زورق من صنع أحلام الشباب
 >. We notice that the word tokenizer generates the most unknowns which is expected given the complexity of the given sentence. The morphological tokenizer creates two unknowns which is lower because it has high chance of adding new words. On the other hand, the stochastic and disjoint letter tokenizers don't create any unknowns because there is a high chance that the learnt vocabulary will have the subword representations. Finally, since SentencePiece is a bottom-up approach, most likely it will not create any unknowns. 
 
\begin{table}[htp!]
\centering
\begin{tabular}{l|l}
\hline
\textbf{Tokenizer} & \textbf{Tokenized form}
\\ \hline
 Word & \< _ في _ من صنع _ الشباب >
 \\ \hline
 SentencePiece & \< ساب +ح في ز +ور +ق من صنع أح +لام الشباب >
 \\ \hline
 Stochastic & \<
 ساب +ح في زو +رق من صن +ع أح +لام الشب +اب
 >
 \\ \hline
 Disjoint Letters & \<
 سا +بح في ز +و +ر +ق من صنع أ +حلا +م ا لشبا +ب
 >
 \\ \hline
 
 Characters & \<
 س +ا +ب +ح ف +ي ز +و +ر +ق م +ن ص +ن +ع أ +ح +ل +ا +م ا +ل +ش +ب +ا +ب
 > 
 \\ \hline
 
 Morphological & \<
 _ في _ من صنع أحلام ال +شباب
 >
 \\ \hline
\end{tabular}
\caption{The output for different tokenizers for an example sentence. $\blacksquare$  represents unknown tokens and + sign represents split token.}
\label{tab:tokenization_examples}
\end{table}

\section{Datasets}
We use four different datasets encompassing three different text classification tasks which are:  sentiment analysis, news classification, and poetry classification.  

\begin{itemize}
    
    \item \textbf{AJGT} The Arabic Jordanian General Tweets dataset contains 1,800 tweets labeled as positive or negative sentiment \citep{alomari2017arabic}. Accordingly, it is a binary classification task. We split the dataset randomly into 1,458 for training, 162 for validation, and 180 for testing. We don't apply any types of preprocessing for this dataset. 
    
    \item  \textbf{Metrec} the dataset contains verses from Arabic poetry along with the meter of the poem \citep{al2020metrec}. The task is to identify the poem meter given a verse of a poem. The dataset consists of poems having 14 different meters. The dataset is split into 47,124 verses for training and 8,316 for testing. A random split of 20\% from the training set is used for validation. The dataset is cleaned by removing diacritics and special characters. 
    
    \item \textbf{LABR} Large-Scale Arabic Book Reviews Dataset contains book reviews \citep{aly2013labr}. We use the dataset for sentiment classification task where we map the 1-star and 2-stars reviews to negative sentiment, 4-stars and 5-stars reviews to positive  sentiment and discard the reviews with rating of 3-stars. The dataset contains 12,502 reviews for training, 658 for validation, and 3,288 for testing.  We don't apply any further preprocessing steps to this dataset. 
    
    \item \textbf{DSAC} DataSet for Arabic Classification. A collection of Arabic texts for newspapers that can be used for text classification \citep{newsdataset}. It contains 111,728 documents annotated with sport, politics, culture, economy, and diverse. We only use a subset of 10K data samples split into 9025, 475, 500 for training, validation, and testing respectively. 

\end{itemize}

In Table \ref{tab:datasets}, we compare the four datasets in terms of the number of words and the number of unique words. We observe that we cover a wide range of datasets ranging from 11.6 thousand to 2.3 million words. We consider AJGT as a low-resource daatset because the number of samples is very low. The table shows the sparsity of the Arabic vocabulary where for AJGT for instance, around 50\% of the words are actually unique.  
\begin{table}[]
\begin{center}
\caption{Number of words and unique words for each dataset}
\label{tab:datasets}
\begin{tabular}{l|c|c|c|c}
\hline
 &\multicolumn{2}{l|}{\textbf{Number of words}} & \multicolumn{2}{l}{\textbf{Number of unique words}} \\ \hline
 & \textbf{Train} & \textbf{Test} & \textbf{Train} & \textbf{Test} \\ \hline
\textbf{AJGT} & 11,689 & 1,486 & 6,453 & 1,101 \\ \hline
\textbf{Metrec} & 341,917 & 75,323 & 102,822 & 42,346 \\ \hline
\textbf{LABR} & 762,758 & 201,531 & 128,673 & 51,524 \\ \hline
\textbf{DSAC} & 2,295,746 & 118,445 & 141,429  & 27,438 \\ \hline

\end{tabular}
\end{center}
\end{table}

\section{Experiments}

In all of our experiments, we use Python for coding and Keras with TensorFlow backend for designing the deep learning models. In order to avoid the bias of random initialization we randomize the training/validation data for each run and also we average three runs for each experiment and report the results. The validation data is used for saving the best model for evaluating the results on the testing data. We use accuracy as the main metric for evaluating our models.

\begin{figure}
    \centering
    \includegraphics[scale=0.09]{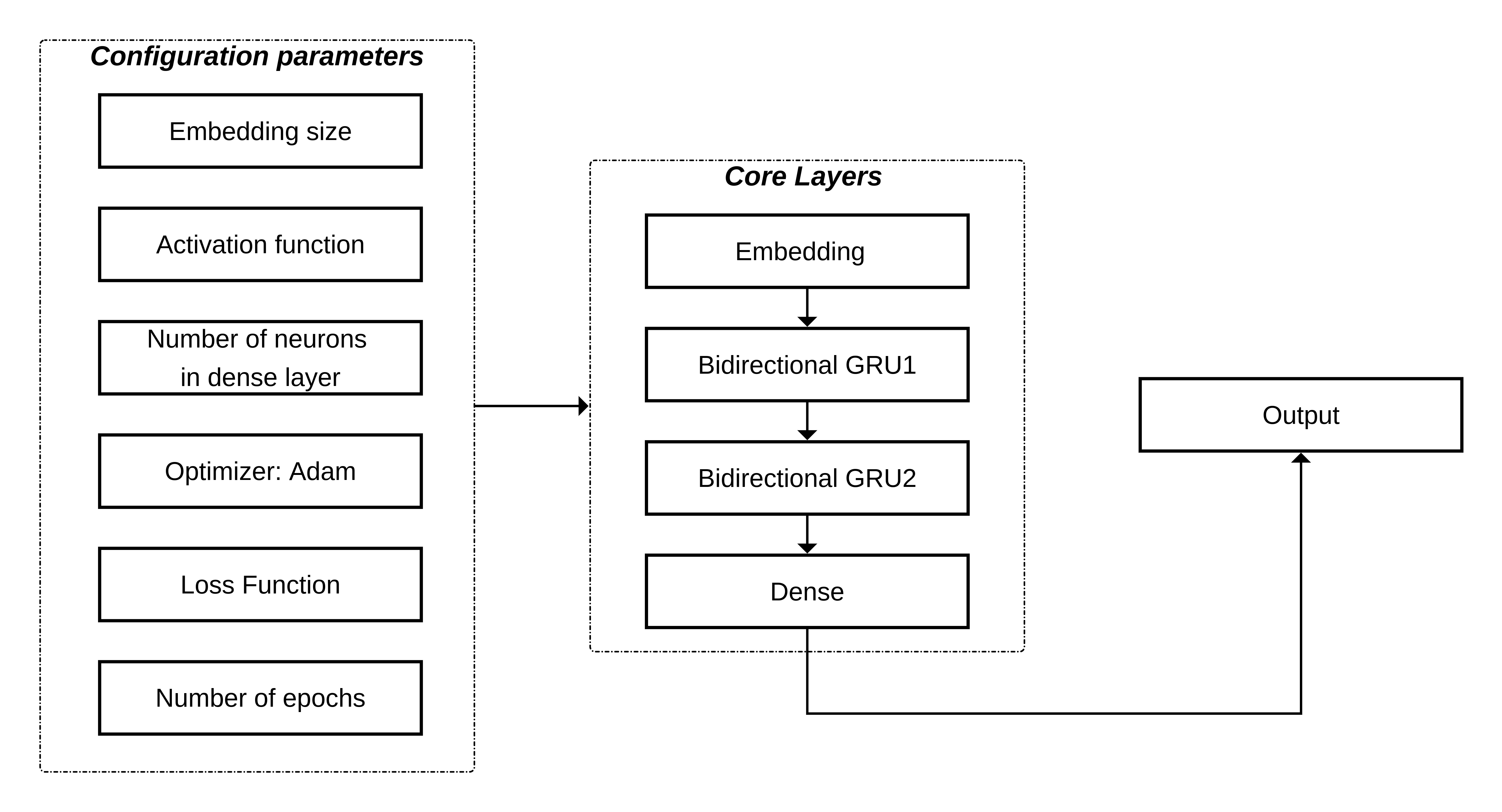}
    \caption{General Architecture.}
    \label{fig:arch}
\end{figure}

\subsection{Experiment setup}
In this subsection, we illustrate the setup we made to conduct the experiment. The goal is to study the behaviour of the tokenizers on different vocabulary sizes. The hypothesis is that the less the vocabulary size, the more critical the tokens are. Also, on the large vocabulary size, the performance should not vary noticeably. We experimented on 500, 1k, 5k, 10k, 20k, and 30k vocabulary sizes. Each vocabulary contains the corresponding tokens to each tokenizer. Furthermore, we run the experiment for each size 3 times and calculated the final accuracy as the average of the runs accuracies.
Figure \ref{fig:arch} illustrates the general architecture we train our experiments on. It consists of an embedding layer followed by two bidirectional GRU layers followed by a fully connected dense layer.

\subsection{Poetry Classification}
We use \textbf{Metrec} for this task. For the model, we train two Bidirectional GRUs with 256 units for each one. We connect that to a dense layer with 14 neurons with Softmax activation. The embedding size is fixed to 128 units. We use sparse categorical cross-entropy as the loss function and Adam as the optimizer. We train the model for 30 epochs and after each epoch, we evaluate it on the validation dataset. We save the model with the minimum validation loss. We use the test set only at the end for the best-saved model with the minimum validation loss.

\subsection{News Classification}
We use \textbf{DSAC} for this task. For the model, we train two Bidirectional GRUs with 256 units for each one. We connect that to a dense layer with 5 neurons and softmax activation. The embedding size is fixed to 128 units. We use cross-entropy as the loss function and Adam as the optimizer. We train the model for 20 epochs and after each epoch, we evaluate it on the validation dataset. We save the model with the minimum validation loss. We use the test set only at the end for the best-saved model with the minimum validation loss. 

\subsection{Sentiment Analysis}
We use \textbf{AJGT} for this task. We consider this as the low resource task with less than 2K training records. For the model, we train two Bidirectional GRUs with 256 units for each one. We connect that to a dense layer with 1 neuron with Sigmoid activation. The embedding size is fixed to 128 units. We use binary cross-entropy as the loss function and Adam as the optimizer. We train the model for 30 epochs and after each epoch, we evaluate it on the validation dataset. We save the model with the minimum validation loss. We use the test set only at the end for the best-saved model with the minimum validation loss. 
We also use the \textbf{LABR} dataset for this task. We consider this as a high-resource task with over 10K training records. For the model, we train two Bidirectional GRUs with 256 units for each one. We connect that to a dense layer with 1 neuron with Sigmoid activation. The embedding size is fixed to 128 units. We use binary cross-entropy as the loss function and Adam as the optimizer. We train the model for 30 epochs and after each epoch, we evaluate it on the validation dataset. We save the model with the minimum validation loss. We use the test set only at the end for the best-saved model with the minimum validation loss.  

\begin{figure}
     \begin{subfigure}[b]{0.3\textwidth}
         \includegraphics[width=\textwidth]{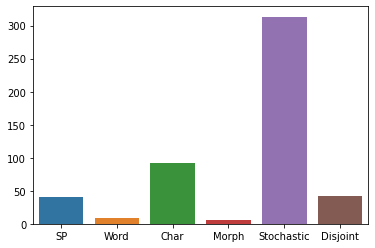}
         \caption{Training speed in seconds.}
     \end{subfigure}
     \hfill
     \begin{subfigure}[b]{0.3\textwidth}
         \includegraphics[width=\textwidth]{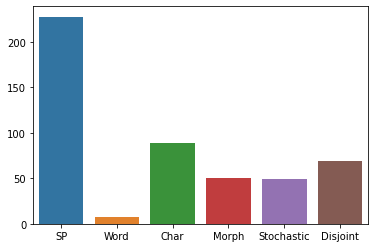}
         \caption{Inference speed in seconds.}
     \end{subfigure}
     \hfill
    \begin{subfigure}[b]{0.3\textwidth}
         \includegraphics[width=\textwidth]{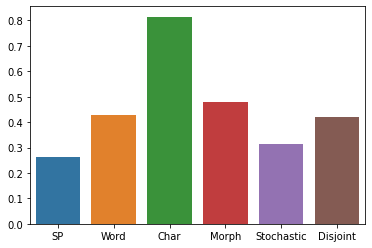}
         \caption{Compression Factor.}
     \end{subfigure}
     \caption{Unsupervised evaluation for the tokenizers.}
        \label{fig:unsupervised}
\end{figure}

\section{Unsupervised Evaluation}
\label{sec:unsupervised}
In this section, we describe how we can evaluate the tokenizers without any fine-tuning. Unsupervised evaluation refers to the evaluation where annotated data is not required. In this evaluation, we introduce two attributes: speed and compression factor. We then conducted an experiment for such evaluation. In this experiment, we  train all the tokenizers on a dataset of size 147 MB extracted from Wikipedia with around 10K vocabulary sizes. 

\subsection{Speed}
There are two main stages for creating a tokenizer, i.e., training and encoding. The output of the training process is a vocabulary that can then be used for tokenization or encoding. Mainly, we can measure the speed of each stage and observe which tokenizer is the best for each one. We conduct all the experiments on a Google Colab virtual machine, hence the results can be reproduced easily. In Figure \ref{fig:unsupervised} we show the training and encoding speed for various tokenizers. We see that training a word and morphology-based tokenization scheme is much faster than other approaches. The stochastic tokenizer has a slower training time because it is trained on all possible segments of a given length chosen randomly. Hence for larger splits, it may taker longer time. For the encoding, we also notice that the word tokenizer achieves the best speed while SentencePiece tokenizer achieves the worse speed. 

\subsection{Compression Factor}
Generally in tokenization, there is a trade-off between creating meaningful tokens and covering as many subwords as we can. For example word tokenizers generate meaningful tokens but it will result in a lot of unknowns while character tokenizers don't generate unknowns but result in non-meaningful character tokens. The idea behind a compression factor is that it takes these two metrics into account. We define the compression factor as :

$$ \text{Compression Factor} = \frac{\sum  \text{total  generated tokens} }{ \sum \text{chars} + \sum \text{words}}$$

We define the number of tokens of the unknown symbol as $\text{len(word)} + 1$. Hence, the compression factor calculates how each tokenizer is able to encode the text into a minimum number of subwords. If the value of compression factor is closer to 1, it means that all the words were encoded as unknowns. On the other hand, if the value is close to 0, it means that all the words are encoded as-is without any splitting. 
From Figure \ref{fig:unsupervised} we see that SentencePiece is the best in terms of encoding using the least number of tokens followed by the stochastic tokenizer. The character tokenizer achieves the worst because it creates a maximum number of tokens for each splitting. Since the morphological tokenizer does pre-segmentation, it will most probably have a lot of unknown symbols and number of tokens, hence it is the worst after the character tokenizer. 
\begin{figure}
     \begin{subfigure}[b]{0.45\textwidth}
         \includegraphics[width=\textwidth]{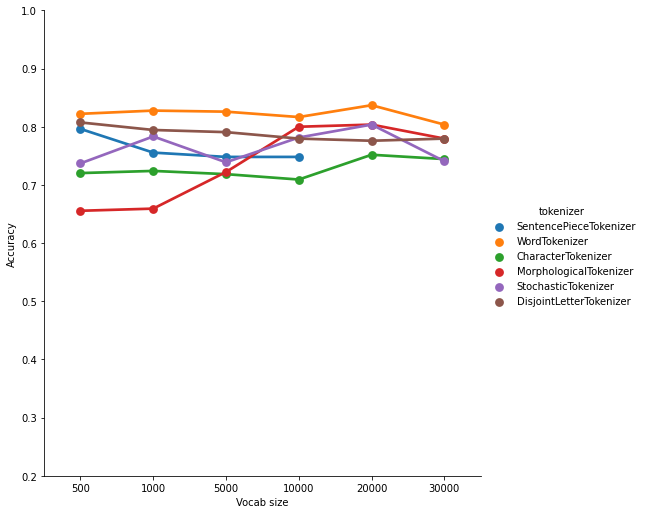}
         \caption{Sentiment Analysis (AJGT)}
     \end{subfigure}
     \hfill
     \begin{subfigure}[b]{0.45\textwidth}
         \includegraphics[width=\textwidth]{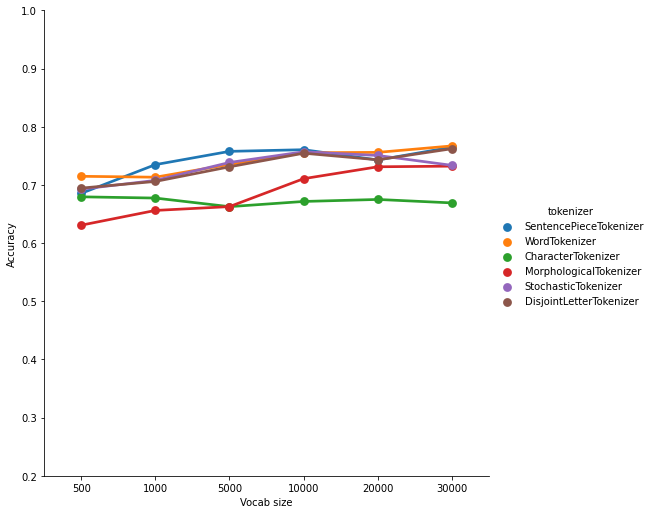}
         \caption{Sentiment Analysis (LABR)}
     \end{subfigure}
     \hfill
    \caption{Results for the two different datasets on the sentiment analysis task. }
    \label{fig:results1}
\end{figure}

\section{Supervised Evaluation}
In this section, we compare the performance, in terms of accuracy, of different tokenizers evaluated on the presented datasets using 6 different vocabulary sizes. Supervised evaluation indicates that the evaluation utilized a labeled dataset. We average the three runs and output the results for each vocabulary size. 

\subsection{Low vs High Resource Tokenization}
In Figure \ref{fig:results1}, we compare the tokenizers for the sentiment analysis task for a fixed number of tokens after tokenization. We consider the first dataset \textbf{AJGT} as a low resource dataset for sentiment analysis while \textbf{LABR} as a high resource dataset for sentiment analysis. For the first dataset, we observe that mostly the word tokenizer achieves the best results among all the tokenizers. This dataset is based on morphology with many words and different dialects which creates a huge data sparsity. Moreover, the maximum number of tokens is 11K which results in the word tokenizer to stop improving massively after a vocabulary size of 10K.  We observe that the morphological tokenizer somewhat achieves reasonable results in addition to the disjoint letters tokenizer. The morphological tokenizer improves a lot after 5K vocabulary size because with smaller vocabulary size it will result in a lot of unkowns. The character tokenizer achieves on average the worse results while the SentencePiece tokenizer stops at 10K because the max number of tokens for the dataset is around 11K. The stochastic tokenizer achieves surprising results and sometimes it is not consistent especially at vocabulary size 10K.
Regarding \textbf{LABR} dataset, the word tokenizer achieves also the best results for larger vocabulary size but for reasonable vocabulary size the disjoint letter tokenizer and SentencePiece tokenizer achieve reasonable results.  The character tokenizer for both tasks achieves the worse results which is expected given the huge data sparsity. The morphological tokenizer follows a similar behaviour to the sentiment analysis task and increases linearly. 
As a comparison between low resource and high resource, we see that for a low resource it is best to use word-based and morphological tokenizers without loss of any linguistic information while for a high resource we see that three tokenizers are actually tied or very closed with vocabulary size 30K. We can conclude that for sentiment analysis, the choice of tokenizer doesn't matter especially if you increase the vocabulary size.

\subsection{Fine-grain vs Coarse grain Tokenization}

As can be analyzed from the sentiment analysis results, coarse-grain tokenization achieves the best on low resource datasets. Subwords tokenization, in such cases, proves to be suboptimal. This can be reasoned as complete word tokens provide more knowledge than subwords on the low resource setup, while subwords tokenization needs more data to reach to a similar level of knowledge. Even SentiencePeice tokenizer completely exhausts the whole low resource \textbf{AJGT} dataset on 10k vocabulary size. This claim is also supported on the high resource results where words tokenization and all subwords tokenizers achieve close results on 30k vocabulary size. Fine-grain character level tokenizer show poor results on both datasts. 

\begin{figure}
     \begin{subfigure}[b]{0.45\textwidth}
         \includegraphics[width=\textwidth]{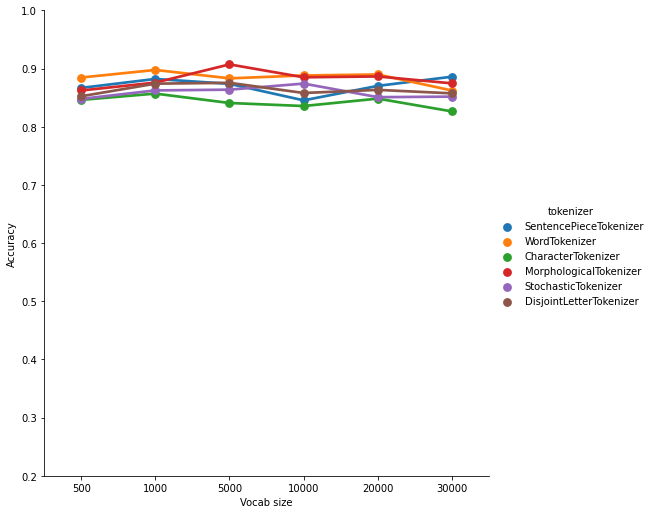}
         \caption{News Classification (DSAC)}
     \end{subfigure}
     \begin{subfigure}[b]{0.45\textwidth}
         \includegraphics[width=\textwidth]{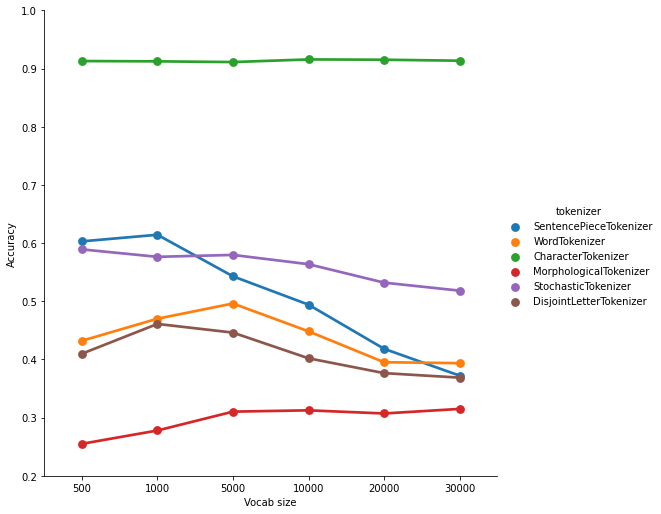}
         \caption{Poetry Classification (MetRec)}
     \end{subfigure}
    \caption{Results for the news classification and poetry classification tasks. }
    \label{fig:results2}
\end{figure}

In Figure \ref{fig:results2}-(b), we observe the results for the poetry classification dataset. It seems that fine-grain tokenization gives the best results on poem-meter classification.  We observe that the character tokenizer, by far, achieves the best results. This is due to the nature of the task which requires character information to classify the poems. For smaller vocabulary sizes, both the stochastic tokenizer and SentencePiece tokenizers achieve decent results because there is a better chance of generating smaller tokens which helps in improving accuracy. We see that the SentencePiece tokenizer degrades quickly because it has a better chance of merging into larger words while the stochastic tokenizer stays at the same level because it uniformly distributes the different possible tokens for a given word.  For other tokenizers, they provide poor results as they are not suitable for this task. However, there is no big variance on large vocabulary sizes where they converge to almost similar results with the exception of the stochastic tokenizer. The behaviour is unpredictable but it achieves average results. Regarding the results on the news classification task as summarized in Figure \ref{fig:results2}-(a), the pattern seems quite noisy due to the variability in data sample sizes. However, on 30k vocabulary size, all tokenizations including stochastic seem to converge to an average of 0.87 accuracy.

\subsection{Morphology vs Plain Data-driven Tokenization}
As we discussed in the previous sections, it is difficult to discriminate between tokenizers especially for datasets that are high resource. Still, it seems capturing morphology improves the accuracy for many datasets like \textbf{AJGT}, \textbf{LABR} and \textbf{DSAC} datasets especially for larger vocabulary size. In Figure \ref{fig:results2}-(a), we compare the tokenizers for the news classification task. The morphological tokenizer achieves better results especially at vocabulary size 5K but it starts degrading after that. The word tokenizer, on the other hand, degrades a lot faster especially after a vocabulary size of 20K because it can't keep up with the complexity of learning more tokens due to the dataset being sparser. The SentencePiece tokenizer achieves surprising results as it degrades at 10K then starts increasing as it learns much better representations. We believe the sudden drop is due to the failure of the model to capture meaningful tokens especially at vocabulary of 10K. This is expected especially due to the huge number of tokens, i.e., around 2 million total tokens with around 140K unique tokens. 

\section{Conclusion}
In this work, we compared six types of tokenizers for three different tasks on Arabic. Mostly, for small vocabulary size, the word tokenizer achieved the best results for sentiment analysis and news classification. The character tokenizer achieved the best results for poetry classification for small and larger vocabulary sizes. We also noticed that increasing the vocabulary size usually results in an improvement in accuracy. Noticeably, the stochastic tokenizer seems to achieve decent results for most tokenizers which indicates that meaningless tokens can achieve good results on Arabic text. We also see the same trend for the disjoint letter tokenizer which seems to get decent results even though the tokens have no meanings in terms of morphology. We believe that very low resource datasets, some information about morphology and the larger the token size the best results but for high resource datasets, it seems that the type of tokenizer doesn't matter as long as you use somewhat coarse-grained tokenization. As future work, it is interesting to see how different tokenizers behave on more complex architectures like attention-based models with different hyper-parameters and complicated tasks like question answering, named entity recognition and translation. This is important especially for tasks that require more meaningful tokens than just a stochastic tokenizer that can give some decent results arbitrarily. 
% talk about choosing different hyperparameters 
\bibliographystyle{plainnat}  
\bibliography{refs}

\begin{thebibliography}{47}
\providecommand{\natexlab}[1]{#1}
\providecommand{\url}[1]{\texttt{#1}}
\expandafter\ifx\csname urlstyle\endcsname\relax
  \providecommand{\doi}[1]{doi: #1}\else
  \providecommand{\doi}{doi: \begingroup \urlstyle{rm}\Url}\fi

\bibitem[Abandah et~al.(2020)Abandah, Khedher, Abdel-Majeed, Mansour, Hulliel,
  and Bisharat]{abandah2020classifying}
Gheith~A Abandah, Mohammed~Z Khedher, Mohammad~R Abdel-Majeed, Hamdi~M Mansour,
  Salma~F Hulliel, and Lara~M Bisharat.
\newblock Classifying and diacritizing arabic poems using deep recurrent neural
  networks.
\newblock \emph{Journal of King Saud University-Computer and Information
  Sciences}, 2020.

\bibitem[Abdelali et~al.(2016)Abdelali, Darwish, Durrani, and
  Mubarak]{abdelali2016farasa}
Ahmed Abdelali, Kareem Darwish, Nadir Durrani, and Hamdy Mubarak.
\newblock Farasa: A fast and furious segmenter for arabic.
\newblock In \emph{Proceedings of the 2016 Conference of the North American
  Chapter of the Association for Computational Linguistics: Demonstrations},
  pages 11--16, 2016.

\bibitem[Abdelali et~al.(2021)Abdelali, Hassan, Mubarak, Darwish, and
  Samih]{abdelali2021pre}
Ahmed Abdelali, Sabit Hassan, Hamdy Mubarak, Kareem Darwish, and Younes Samih.
\newblock Pre-training bert on arabic tweets: Practical considerations.
\newblock \emph{arXiv preprint arXiv:2102.10684}, 2021.

\bibitem[Abdul-Mageed et~al.(2020)Abdul-Mageed, Elmadany, and
  Nagoudi]{abdul2020arbert}
Muhammad Abdul-Mageed, AbdelRahim Elmadany, and El~Moatez~Billah Nagoudi.
\newblock Arbert \& marbert: deep bidirectional transformers for arabic.
\newblock \emph{arXiv preprint arXiv:2101.01785}, 2020.

\bibitem[{Abu Farha} and Magdy(2021)]{ABUFARHA2021102438}
Ibrahim {Abu Farha} and Walid Magdy.
\newblock A comparative study of effective approaches for arabic sentiment
  analysis.
\newblock \emph{Information Processing \& Management}, 58\penalty0
  (2):\penalty0 102438, 2021.
\newblock ISSN 0306-4573.
\newblock \doi{https://doi.org/10.1016/j.ipm.2020.102438}.
\newblock URL
  \url{http://www.sciencedirect.com/science/article/pii/S0306457320309316}.

\bibitem[Al-Ayyoub et~al.(2019)Al-Ayyoub, Khamaiseh, Jararweh, and
  Al-Kabi]{al2019comprehensive}
Mahmoud Al-Ayyoub, Abed~Allah Khamaiseh, Yaser Jararweh, and Mohammed~N
  Al-Kabi.
\newblock A comprehensive survey of arabic sentiment analysis.
\newblock \emph{Information processing \& management}, 56\penalty0
  (2):\penalty0 320--342, 2019.

\bibitem[Al-Helali and Mahmoud(2017)]{al2017arabic}
Baligh~M Al-Helali and Sabri~A Mahmoud.
\newblock Arabic online handwriting recognition (aohr) a survey.
\newblock \emph{ACM Computing Surveys (CSUR)}, 50\penalty0 (3):\penalty0 1--35,
  2017.

\bibitem[Al-Rfou et~al.(2019)Al-Rfou, Choe, Constant, Guo, and
  Jones]{al2019character}
Rami Al-Rfou, Dokook Choe, Noah Constant, Mandy Guo, and Llion Jones.
\newblock Character-level language modeling with deeper self-attention.
\newblock In \emph{Proceedings of the AAAI Conference on Artificial
  Intelligence}, volume~33, pages 3159--3166, 2019.

\bibitem[Al-shaibani et~al.(2020)Al-shaibani, Alyafeai, and Ahmad]{al2020meter}
Maged~S Al-shaibani, Zaid Alyafeai, and Irfan Ahmad.
\newblock Meter classification of arabic poems using deep bidirectional
  recurrent neural networks.
\newblock \emph{Pattern Recognition Letters}, 2020.

\bibitem[Al-Shaibani et~al.(2020)Al-Shaibani, Alyafeai, and
  Ahmad]{al2020metrec}
Maged~S Al-Shaibani, Zaid Alyafeai, and Irfan Ahmad.
\newblock Metrec: A dataset for meter classification of arabic poetry.
\newblock \emph{Data in Brief}, 33:\penalty0 106497, 2020.

\bibitem[Alkaoud and Syed(2020)]{alkaoud2020importance}
Mohamed Alkaoud and Mairaj Syed.
\newblock On the importance of tokenization in arabic embedding models.
\newblock In \emph{Proceedings of the Fifth Arabic Natural Language Processing
  Workshop}, pages 119--129, 2020.

\bibitem[Alomari et~al.(2017)Alomari, ElSherif, and Shaalan]{alomari2017arabic}
Khaled~Mohammad Alomari, Hatem~M ElSherif, and Khaled Shaalan.
\newblock Arabic tweets sentimental analysis using machine learning.
\newblock In \emph{International Conference on Industrial, Engineering and
  Other Applications of Applied Intelligent Systems}, pages 602--610. Springer,
  2017.

\bibitem[Aly and Atiya(2013)]{aly2013labr}
Mohamed Aly and Amir Atiya.
\newblock Labr: A large scale arabic book reviews dataset.
\newblock In \emph{Proceedings of the 51st Annual Meeting of the Association
  for Computational Linguistics (Volume 2: Short Papers)}, pages 494--498,
  2013.

\bibitem[Antoun et~al.(2020)Antoun, Baly, and Hajj]{antoun2020arabert}
Wissam Antoun, Fady Baly, and Hazem Hajj.
\newblock Arabert: Transformer-based model for arabic language understanding.
\newblock \emph{arXiv preprint arXiv:2003.00104}, 2020.

\bibitem[Atallah and Omar(2009)]{atallah2009comparative}
AL-Shatnawi Atallah and Khairuddin Omar.
\newblock A comparative study between methods of arabic baseline detection.
\newblock In \emph{2009 International Conference on Electrical Engineering and
  Informatics}, volume~1, pages 73--77. IEEE, 2009.

\bibitem[Badaro et~al.(2019)Badaro, Baly, Hajj, El-Hajj, Shaban, Habash,
  Al-Sallab, and Hamdi]{badaro2019survey}
Gilbert Badaro, Ramy Baly, Hazem Hajj, Wassim El-Hajj, Khaled~Bashir Shaban,
  Nizar Habash, Ahmad Al-Sallab, and Ali Hamdi.
\newblock A survey of opinion mining in arabic: a comprehensive system
  perspective covering challenges and advances in tools, resources, models,
  applications, and visualizations.
\newblock \emph{ACM Transactions on Asian and Low-Resource Language Information
  Processing (TALLIP)}, 18\penalty0 (3):\penalty0 1--52, 2019.

\bibitem[BINIZ(2018)]{newsdataset}
mohamed BINIZ.
\newblock Dataset for arabic classification.
\newblock 2018.

\bibitem[Bojanowski et~al.(2017)Bojanowski, Grave, Joulin, and
  Mikolov]{bojanowski2017enriching}
Piotr Bojanowski, Edouard Grave, Armand Joulin, and Tomas Mikolov.
\newblock Enriching word vectors with subword information.
\newblock \emph{Transactions of the Association for Computational Linguistics},
  5:\penalty0 135--146, 2017.
\newblock ISSN 2307-387X.

\bibitem[Bostrom and Durrett(2020)]{bostrom2020byte}
Kaj Bostrom and Greg Durrett.
\newblock Byte pair encoding is suboptimal for language model pretraining.
\newblock \emph{arXiv preprint arXiv:2004.03720}, 2020.

\bibitem[Chitnis and DeNero(2015)]{chitnis2015variable}
Rohan Chitnis and John DeNero.
\newblock Variable-length word encodings for neural translation models.
\newblock In \emph{Proceedings of the 2015 Conference on Empirical Methods in
  Natural Language Processing}, pages 2088--2093, 2015.

\bibitem[Devlin et~al.(2018)Devlin, Chang, Lee, and Toutanova]{devlin2018bert}
Jacob Devlin, Ming-Wei Chang, Kenton Lee, and Kristina Toutanova.
\newblock Bert: Pre-training of deep bidirectional transformers for language
  understanding.
\newblock \emph{arXiv preprint arXiv:1810.04805}, 2018.

\bibitem[El-Khair(2016)]{el20161}
Ibrahim~Abu El-Khair.
\newblock 1.5 billion words arabic corpus.
\newblock \emph{arXiv preprint arXiv:1611.04033}, 2016.

\bibitem[ElJundi et~al.(2019)ElJundi, Antoun, El~Droubi, Hajj, El-Hajj, and
  Shaban]{eljundi2019hulmona}
Obeida ElJundi, Wissam Antoun, Nour El~Droubi, Hazem Hajj, Wassim El-Hajj, and
  Khaled Shaban.
\newblock hulmona: The universal language model in arabic.
\newblock In \emph{Proceedings of the Fourth Arabic Natural Language Processing
  Workshop}, pages 68--77, 2019.

\bibitem[Guellil et~al.(2019)Guellil, Azouaou, and Mendoza]{guellil2019arabic}
Imane Guellil, Faical Azouaou, and Marcelo Mendoza.
\newblock Arabic sentiment analysis: studies, resources, and tools.
\newblock \emph{Social Network Analysis and Mining}, 9\penalty0 (1):\penalty0
  56, 2019.

\bibitem[Howard and Ruder(2018)]{howard2018universal}
Jeremy Howard and Sebastian Ruder.
\newblock Universal language model fine-tuning for text classification.
\newblock \emph{arXiv preprint arXiv:1801.06146}, 2018.

\bibitem[Jasim(2020)]{jasimarabic}
Mahdi~Nsaif Jasim.
\newblock Arabic optical characters recognition by neural network based arabic
  unicode.
\newblock 2020.

\bibitem[Kudo(2018)]{kudo2018subword}
Taku Kudo.
\newblock Subword regularization: Improving neural network translation models
  with multiple subword candidates.
\newblock \emph{arXiv preprint arXiv:1804.10959}, 2018.

\bibitem[Kudo and Richardson(2018)]{kudo2018sentencepiece}
Taku Kudo and John Richardson.
\newblock Sentencepiece: A simple and language independent subword tokenizer
  and detokenizer for neural text processing.
\newblock \emph{arXiv preprint arXiv:1808.06226}, 2018.

\bibitem[Kunchukuttan and Bhattacharyya(2016)]{kunchukuttan2016orthographic}
Anoop Kunchukuttan and Pushpak Bhattacharyya.
\newblock Orthographic syllable as basic unit for smt between related
  languages.
\newblock \emph{arXiv preprint arXiv:1610.00634}, 2016.

\bibitem[Liu et~al.(2019)Liu, Ott, Goyal, Du, Joshi, Chen, Levy, Lewis,
  Zettlemoyer, and Stoyanov]{liu2019roberta}
Yinhan Liu, Myle Ott, Naman Goyal, Jingfei Du, Mandar Joshi, Danqi Chen, Omer
  Levy, Mike Lewis, Luke Zettlemoyer, and Veselin Stoyanov.
\newblock Roberta: A robustly optimized bert pretraining approach.
\newblock \emph{arXiv preprint arXiv:1907.11692}, 2019.

\bibitem[Mikolov et~al.(2013)Mikolov, Chen, Corrado, and
  Dean]{mikolov2013efficient}
Tomas Mikolov, Kai Chen, Greg Corrado, and Jeffrey Dean.
\newblock Efficient estimation of word representations in vector space.
\newblock \emph{arXiv preprint arXiv:1301.3781}, 2013.

\bibitem[Oudah et~al.(2019)Oudah, Almahairi, and Habash]{oudah2019impact}
Mai Oudah, Amjad Almahairi, and Nizar Habash.
\newblock The impact of preprocessing on arabic-english statistical and neural
  machine translation.
\newblock \emph{arXiv preprint arXiv:1906.11751}, 2019.

\bibitem[Pasha et~al.(2014)Pasha, Al-Badrashiny, Diab, El~Kholy, Eskander,
  Habash, Pooleery, Rambow, and Roth]{pasha2014madamira}
Arfath Pasha, Mohamed Al-Badrashiny, Mona~T Diab, Ahmed El~Kholy, Ramy
  Eskander, Nizar Habash, Manoj Pooleery, Owen Rambow, and Ryan Roth.
\newblock Madamira: A fast, comprehensive tool for morphological analysis and
  disambiguation of arabic.
\newblock In \emph{LREC}, volume~14, pages 1094--1101, 2014.

\bibitem[Pennington et~al.(2014)Pennington, Socher, and
  Manning]{pennington2014glove}
Jeffrey Pennington, Richard Socher, and Christopher~D Manning.
\newblock Glove: Global vectors for word representation.
\newblock In \emph{Proceedings of the 2014 conference on empirical methods in
  natural language processing (EMNLP)}, pages 1532--1543, 2014.

\bibitem[Radford et~al.(2019)Radford, Wu, Child, Luan, Amodei, and
  Sutskever]{radford2019language}
Alec Radford, Jeffrey Wu, Rewon Child, David Luan, Dario Amodei, and Ilya
  Sutskever.
\newblock Language models are unsupervised multitask learners.
\newblock \emph{OpenAI Blog}, 1\penalty0 (8):\penalty0 9, 2019.

\bibitem[Raffel et~al.(2019)Raffel, Shazeer, Roberts, Lee, Narang, Matena,
  Zhou, Li, and Liu]{raffel2019exploring}
Colin Raffel, Noam Shazeer, Adam Roberts, Katherine Lee, Sharan Narang, Michael
  Matena, Yanqi Zhou, Wei Li, and Peter~J Liu.
\newblock Exploring the limits of transfer learning with a unified text-to-text
  transformer.
\newblock \emph{arXiv preprint arXiv:1910.10683}, 2019.

\bibitem[Sajjad et~al.(2017)Sajjad, Dalvi, Durrani, Abdelali, Belinkov, and
  Vogel]{sajjad2017challenging}
Hassan Sajjad, Fahim Dalvi, Nadir Durrani, Ahmed Abdelali, Yonatan Belinkov,
  and Stephan Vogel.
\newblock Challenging language-dependent segmentation for arabic: An
  application to machine translation and part-of-speech tagging.
\newblock \emph{arXiv preprint arXiv:1709.00616}, 2017.

\bibitem[Schuster and Nakajima(2012)]{schuster2012japanese}
Mike Schuster and Kaisuke Nakajima.
\newblock Japanese and korean voice search.
\newblock In \emph{2012 IEEE International Conference on Acoustics, Speech and
  Signal Processing (ICASSP)}, pages 5149--5152. IEEE, 2012.

\bibitem[Sennrich et~al.(2015)Sennrich, Haddow, and Birch]{sennrich2015neural}
Rico Sennrich, Barry Haddow, and Alexandra Birch.
\newblock Neural machine translation of rare words with subword units.
\newblock \emph{arXiv preprint arXiv:1508.07909}, 2015.

\bibitem[Shapiro and Duh(2018{\natexlab{a}})]{shapiro2018bpe}
Pamela Shapiro and Kevin Duh.
\newblock Bpe and charcnns for translation of morphology: A cross-lingual
  comparison and analysis.
\newblock \emph{arXiv preprint arXiv:1809.01301}, 2018{\natexlab{a}}.

\bibitem[Shapiro and Duh(2018{\natexlab{b}})]{shapiro2018morphological}
Pamela Shapiro and Kevin Duh.
\newblock Morphological word embeddings for arabic neural machine translation
  in low-resource settings.
\newblock In \emph{Proceedings of the Second Workshop on Subword/Character
  LEvel Models}, pages 1--11, 2018{\natexlab{b}}.

\bibitem[Smit et~al.(2014)Smit, Virpioja, Grönroos, and
  Kurimo]{SmitVirpiojaGronroosEtAl2014}
Peter Smit, Sami Virpioja, Stig-Arne Grönroos, and Mikko Kurimo.
\newblock Morfessor 2.0: Toolkit for statistical morphological segmentation.
\newblock page~4. Aalto University, 2014.
\newblock URL \url{http://urn.fi/URN:NBN:fi:aalto-201409292677}.

\bibitem[Tay et~al.(2021)Tay, Tran, Ruder, Gupta, Chung, Bahri, Qin,
  Baumgartner, Yu, and Metzler]{tay2021charformer}
Yi~Tay, Vinh~Q Tran, Sebastian Ruder, Jai Gupta, Hyung~Won Chung, Dara Bahri,
  Zhen Qin, Simon Baumgartner, Cong Yu, and Donald Metzler.
\newblock Charformer: Fast character transformers via gradient-based subword
  tokenization.
\newblock \emph{arXiv preprint arXiv:2106.12672}, 2021.

\bibitem[Wang et~al.(2021)Wang, Ruder, and Neubig]{wang2021multi}
Xinyi Wang, Sebastian Ruder, and Graham Neubig.
\newblock Multi-view subword regularization.
\newblock \emph{arXiv preprint arXiv:2103.08490}, 2021.

\bibitem[Wu et~al.(2016)Wu, Schuster, Chen, Le, Norouzi, Macherey, Krikun, Cao,
  Gao, Macherey, et~al.]{wu2016google}
Yonghui Wu, Mike Schuster, Zhifeng Chen, Quoc~V Le, Mohammad Norouzi, Wolfgang
  Macherey, Maxim Krikun, Yuan Cao, Qin Gao, Klaus Macherey, et~al.
\newblock Google's neural machine translation system: Bridging the gap between
  human and machine translation.
\newblock \emph{arXiv preprint arXiv:1609.08144}, 2016.

\bibitem[Xue et~al.(2021)Xue, Barua, Constant, Al-Rfou, Narang, Kale, Roberts,
  and Raffel]{xue2021byt5}
Linting Xue, Aditya Barua, Noah Constant, Rami Al-Rfou, Sharan Narang, Mihir
  Kale, Adam Roberts, and Colin Raffel.
\newblock Byt5: Towards a token-free future with pre-trained byte-to-byte
  models.
\newblock \emph{arXiv preprint arXiv:2105.13626}, 2021.

\bibitem[Yousef et~al.(2019)Yousef, Ibrahime, Madbouly, and
  Mahmoud]{yousef2019learning}
Waleed~A Yousef, Omar~M Ibrahime, Taha~M Madbouly, and Moustafa~A Mahmoud.
\newblock Learning meters of arabic and english poems with recurrent neural
  networks: a step forward for language understanding and synthesis.
\newblock \emph{arXiv preprint arXiv:1905.05700}, 2019.

\end{thebibliography}
\end{document}